\documentclass{Interspeech2024}
\interspeechcameraready 

\title{Sustainable self-supervised learning for speech representations}

\name{Luis}{Lugo}
\name{Valentin}{Vielzeuf}
\address{Orange, Cesson-Sévigné, France}
\email{luiseduardo.lugomartinez@orange.com, valentin.vielzeuf@orange.com}


\begin{document}
\maketitle

\begin{abstract}
Sustainable artificial intelligence focuses on data, hardware, and algorithms to make machine learning models more environmentally responsible. In particular, machine learning models for speech representations are computationally expensive, generating environmental concerns because of their high energy consumption. Thus, we propose a sustainable self-supervised model to learn speech representation, combining optimizations in neural layers and training to reduce computing costs. The proposed model improves over a resource-efficient baseline, reducing both memory usage and computing cost estimations. It pretrains using a single GPU in less than a day. On top of that, it improves the error rate performance of the baseline in downstream task evaluations. When comparing it to large speech representation approaches, there is an order of magnitude reduction in memory usage, while computing cost reductions represent almost three orders of magnitude improvement.
\end{abstract}

\section{Introduction}
\label{sec:introduction}

The sustainable development of artificial intelligence requires environmentally responsible approaches, improving models' performance while reducing their computational resource needs \cite{wu2022sustainable}. An area of research with considerable requirements in this regard is self-supervising learning \cite{mohamed2022self,lugo2023efficiency}. Indeed, self-supervised models for speech representation learning are computationally expensive, requiring tens or even hundreds of GPUs during days of pretraining \cite{chen2023reducing,lugo2023efficiency,zhang2020pushing}.

These large computational costs limit the number of laboratories able to study self-supervised approaches, as few research teams have access to large computing resources. A similar situation arises when trying to replicate results with constrained computing resources \cite{chen2023reducing,lin2022melhubert}. Computational costs also pose problems for personalized self-supervised pretraining or model finetuning in mobile devices. The same scenario happens when training models in edge platforms, where only a GPU is available \cite{gaol2023match}. In federated learning, where there can be restrictions to centralize data, approaches train the models using several workers, each worker computing gradients for its own data. When workers are mobile devices or edge platforms, large computing requirements represent a serious limitation \cite{gaol2023match, kairouz2021advances}. Additionally, large computing costs represent environmental concerns because of the high energy consumption they generate. 

We therefore propose a sustainable approach to learn speech representations in a self-supervised way. The proposed model requires only one GPU to pretrain the neural architecture in less than 24 hours. It improves over a resource-efficient baseline in error rate, computing costs, and memory usage.

\section{Related work}
\label{sec:background}

Self-supervised learning is an approach that trains neural models without labeled datasets. First, models are pretrained to learn a latent representation. Then, they can be fine-tuned for a downstream task like phoneme recognition (PR), intent classification (IC), or automatic speech recognition (ASR), in a supervised way \cite{chen2020simple,huang2022spiral,zhang2020pushing}.

Self-supervised models can use a contrastive or predictive loss to learn speech representations \cite{lugo2024towards}. A popular approach using a predictive loss is Hidden Unit BERT (HuBERT) \cite{hsu2021hubert}. It uses masked segments of input speech to pretrain the model. The pretraining comprises two stages. The first stage pretrains the model with kmeans on Mel-frequency cepstral coefficients. The second stage pretrains the model with kmeans assignments from the representation learned in the first stage. After pretraining, a finetuning stage trains the model for a downstream task\textemdash like ASR, for example. Similar to HuBERT, DinoSR \cite{liu2023dinosr} combines masked input pretraining and online clustering for target discretization. Other methods in the predictive loss category include WavLM \cite{chen2022wavlm} and data2vec2 \cite{baevski2023efficient}, which supports image, text, and speech inputs. 

On the other hand, wav2vec2 \cite{baevski2020wav2vec} is a popular approach in the contrastive loss category. The training relies on masked input timesteps and quantization\textemdash through Gumbel SoftMax \cite{jangcategorical}\textemdash to learn latent speech representations. Self-supervised perturbation invariant representation learning (SPIRAL) also pertains to the contrastive loss category. But in contrast to wav2vec2, it uses no quantization and reduces training costs \cite{huang2022spiral}. Combined SSL (ComSSL) replaces the original raw waveforms in wav2vec2 with Mel-frequency cepstral coefficients, which make the input sequence shorter. It also removes the quantization module, using a linear layer instead. ComSSL generates the lowest error rate in ASR \cite{zhang2020pushing}.

\begin{figure*}[htb]
    \centering
    \includegraphics[width=0.80\textwidth]{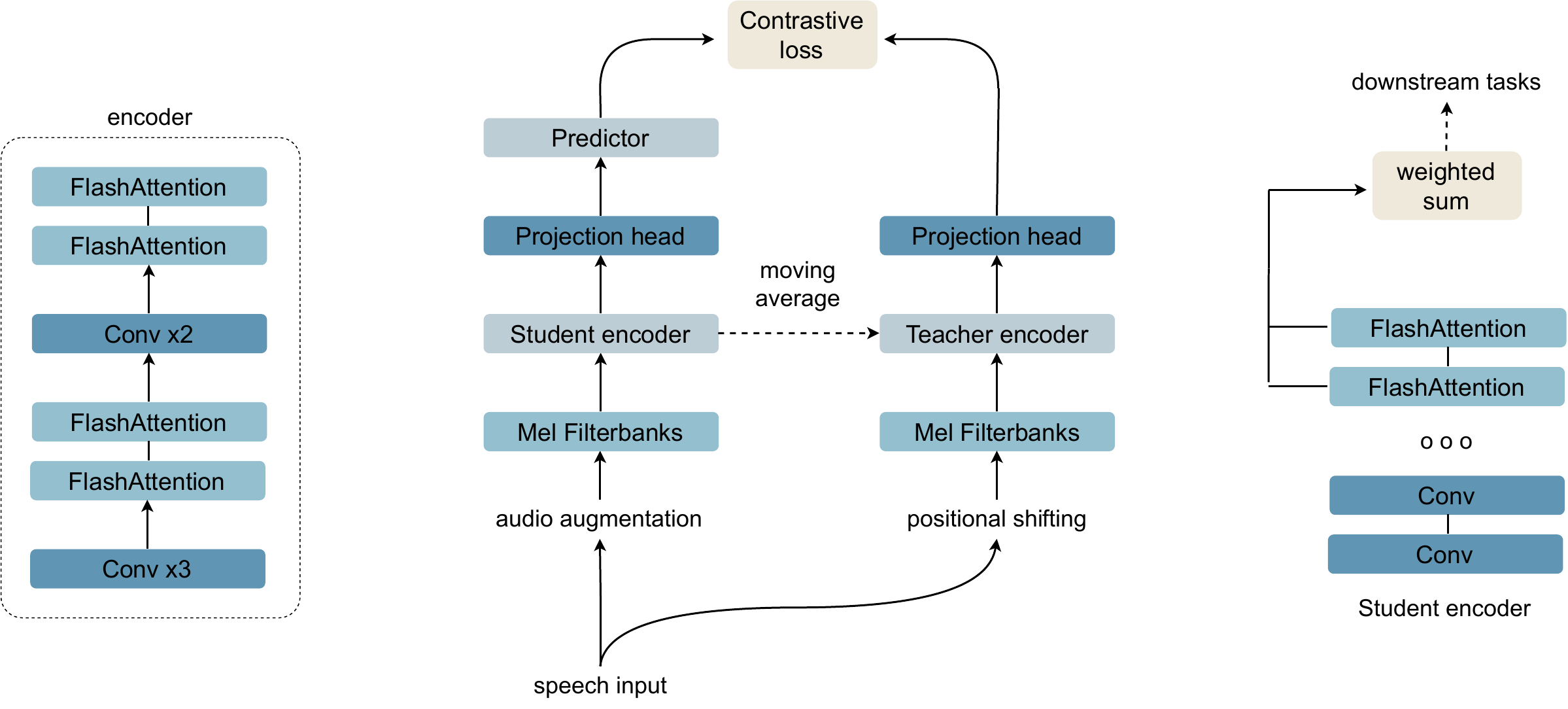}
    \caption{Neural architecture for the S3LSpeech model. It relies on a teacher\textendash student configuration during pretraining \cite{liu2023dinosr}. Both student and teacher encoders have the same layer configuration, comprising interleaved convolutional and FlashAttention layers.}
    \label{fig:s3lspeech_architecture}
\end{figure*}

But regardless of remarkable results in speech processing, self-supervised models for speech representation learning are computationally expensive \cite{lugo2023efficiency,mohamed2022self}. HuBERT, for example, requires 32 GPUs during more than 60 hours to pretrain the model, which contains 95M of trainable parameters in its base version \cite{hsu2021hubert}. Similarly, ComSSL requires 512 Tensor Processing Units (TPUs) for four days to train a model containing 1000M trainable parameters \cite{zhang2020pushing}.

Several approaches have been looking to address large computing costs. Some methods use knowledge distillation \cite{ashihara2022deep,chang2022distilhubert,lee2022fithubert}, pruning \cite{lai2021parp,peng2023structured}, or both \cite{peng2023dphubert}. Efficient self-supervised learning (ESSL) establishes a limit on computational resources for pretraining, using one GPU for 24 to 48 hours of training \cite{ geiping2023cramming,lugo2024towards}. MelHuBERT removes the convolutional module for feature extraction in HuBERT, replacing it with Mel Spectrograms, and reducing the amount of iterative training stages \cite{lin2022melhubert,parcollet2023efficiency}. There are also efforts to squeeze wav2vec2 to make it more efficient \cite{vyas2022demand,wu2022performance}, or trim input speech sequences to be able to train wav2vec2 with limited memory \cite{gaol2023match}. Other approaches replace standard self-attention layers with alternative layers to reduce complexity \cite{parcollet2023sumformer}. 

Despite previous work, existing efforts to improve self-supervised model efficiency have several limitations. Approaches relying on knowledge distillation or pruning still require a full pretrained model, as they cannot be trained standalone \cite{chen2023reducing}. This requirement forces computing cost estimations to include pretraining costs for the full model as well. Moreover, most methods reducing computing costs \cite{chen2023reducing, huang2022spiral, liu2023dinosr} need several GPUs for days of pretraining\textemdash even MelHuBERT, which uses one GPU, requires more than six days of pretraining \cite{lin2022melhubert}. Besides, some approaches fail to use inner self-attention layer outputs for downstream tasks, even though each inner layer learns to encode key aspects of the speech waveform \cite{pasad2023comparative}. Thus, using only the last layer output, or a projection of it \cite{huang2022spiral,lugo2024towards}, might waste key features learned in the inner layers of the representation encoder. 

\section{Sustainable approach for speech representations}
\label{sec:method}

We thus propose Sustainable Self-Supervised Learning for Speech representations (S3LSpeech) to address previously mentioned limitations.

S3LSpeech relies on a teacher\textendash student configuration (Figure \ref{fig:s3lspeech_architecture}), where the student encoder learns to match the representation generated from the teacher encoder for a given speech input \cite{baevski2020wav2vec,liu2023dinosr}. The encoder architecture for both teacher and student combines convolutional and self-attention layers to process the input speech (Table \ref{tab:model_architecture}). The input sequence is first preprocessed to extract a Mel spectrogram, then sent to the first convolutional block \cite{lin2022melhubert}, one of the most efficient configurations for speech preprocessing \cite{parcollet2023efficiency}. 

After the convolutional block, self-attention layers process the input. Standard self-attention layers are inefficient because their time and memory complexity are quadratic in sequence length \cite{dao2022flashattention,parcollet2023sumformer}. Instead, S3LSpeech uses FlashAttention \cite{dao2022flashattention}, an optimized self-attention layer. FlashAttention optimizes self-attention calculations by reducing the read and write operations between the high bandwidth memory and the SRAM in the GPU, without doing any approximations. 

While student weights are trained using backpropagation, teacher weights are updated by tracking the student weights with an exponential moving average (EMA). Formally, EMA is defined as $\theta_t \leftarrow \alpha \theta_t + (1 - \alpha) \theta_s$, where $\theta_t$ are the teacher weights, $\theta_s$ are the student weights, and $\alpha$ is a constant for the convex combination of teacher and student weights during pretraining, controlling the decay rate of the teacher model in each pretraining step \cite{liu2023dinosr}.  

During pretraining, we use a contrastive loss to maximize the agreement between the representations that student and teacher encoders calculate. Formally, the contrastive loss is defined as follows \cite{chen2020simple,huang2022spiral}:

\begin{equation}
  \phi (a, b) = \frac{a^T b}{\left\lVert a \right\rVert \left\lVert b \right\rVert} 
\end{equation}
\begin{equation}
  \mathcal{L} = - \sum_{i = 1}^{T} \log \frac{e^{\phi(z_i, z'_i)/\tau}}{\sum_{j \in D_i}e^{\phi(z_i, z'_j)/\tau}} 
\end{equation}

where $z$ is the representation from the student encoder, $z'$ is the representation from the teacher encoder, $\tau$ is a temperature parameter, and $D_i$ is the audio agumentation for the $z_i$ input.

Though the speech input is the same for both encoders, the student encoder takes as input the augmented sample of the input sequence. Augmented samples avoid a trivial solution, where the student's weights equal the teacher's weights, so we add two audio augmentations to the speech input. First, we add noise under a multicondition pretraining setup \cite{chiba2019multi}. Secondly, we use masks in the time and frequency domain of the input spectrogram, an approach commonly known as SpecAugment \cite{park2019specaugment}. In SpecAugment, randomly selected masks are set to zero in the time domain, while randomly selected masks in the frequency domain are filled out with Gaussian noise \cite{huang2022spiral}.

Along with multicondition pretrain and SpecAugment, we add random positional shifting \cite{huang2022spiral}, a training technique based on randomly shifting the teacher's input sequence in the time domain. This shifting technique prevents the teacher\textendash student architecture from learning any positional information, forcing the student encoder to learn exclusively from speech information in the input sequence \cite{huang2022spiral}.

\begin{table}[bht]
\centering
\begin{tabular}{ p{3.0cm} m{2.5cm}  m{1.0cm} }
    \toprule
    Module  & Configuration & Layers  \\
    \midrule
    \midrule
    Encoder  & & \\
    - Convolutional    & \{ 384, 5, 2 \}     & 1 \\
    - Convolutional    & \{ 512, 5, 2 \}     & 1 \\
    - Convolutional    & \{ 512, 1, 1 \}     & 1 \\ 
    - FlashAttention   & \{ 512, 2048, 8 \}  & 2\\ 
    - Convolutional    & \{ 1536, 5, 2 \}    & 1 \\
    - Convolutional    & \{ 768, 1, 1 \}     & 1 \\
    - FlashAttention   & \{ 512, 3072, 12 \} & 2 \\  
    \midrule
    Projection head & & \\
    - Linear             & \{ 256 \} & 1 \\
    \midrule
    Predictor & & \\
    - Convolutional      & \{ 256, 5, 1 \} & 2 \\
    - Convolutional      & \{ 256, 1, 1 \} & 1 \\
    \bottomrule
    \end{tabular}
    \caption{Layer configurations for the S3LSpeech architecture: convolutional layers \{kernel size, channels, stride\}, linear layers \{size\}, and  FlashAttention layers \{embedded dimension, feedforward embedded dimension, attention heads\}.}
    \label{tab:model_architecture}
\end{table}

We use dynamic batching to maximize the amount of speech in each training batch \cite{ravanelli2021speechbrain,tyagi2020taming}. We also accumulate gradients for a few training steps to get close to batch sizes of large speech models \cite{huang2023measuring}. Batch sizes are crucial for optimizing contrastive losses in self-supervised learning \cite{chen2023reducing,lugo2024towards}, so it is essential to get as close as possible to batch sizes of existing models. Dynamic batching and gradient accumulation enable an effective batch size close to existing speech models, but training with only one GPU.

Both pretraining and finetuning are performed with mixed precision training, which reduces the size of floating point numbers from 32 to 16 bits, without affecting training convergence \cite{micikevicius2018mixed,narayanan2021efficient}. Once pretraining finishes, a weighted sum of the top FlashAttention layers becomes the input to the downstream task module (Figure \ref{fig:s3lspeech_architecture}). The model is then finetuned according to the downstream task \cite{wang2023read}. 

\section{Experimental results}
\label{sec:results}

In this section, we discuss the experimental setup and the results from the S3LSpeech approach. Code is publicly available\footnote{\url{https://github.com/Orange-OpenSource/s3lspeech}} to share experimental settings and facilitate results reproducibility.

\subsection{Experimental setup}

We pretrain the model with 960h of LibriSpeech \cite{panayotov2015librispeech} for 12500 iterations, which represents 50000 iterations with 4 gradient accumulations. During pretraining, we train with a learning rate of 3e-4 with a cosine schedule. For finetuning, we train for 60000 iterations with 100h of Librispeech. Finetuning uses a learning rate of 3e-4, with a warmup of 10\%, 80\% at maximum learning rate, and exponential decay for the remaining 10\% of training iterations. For very-low data experiments, Librilight \cite{kahn2020libri} has 10 hours, 1 hour, and 10 minutes datasets to finetune the model. Additionally, we use the Student's paired t-test (scipy.stats.ttest\_ind) for statistical significance calculations.

As a baseline, we select ESSL \cite{lugo2024towards}, a recent model for speech representation learning. As mentioned (Section \ref{sec:background}), ESSL limits the computational resources available for pretraining the neural model, fixing a limit of 24 to 48 hours of training using a single GPU. For a fair comparison against the baseline, we use the NVIDIA GeForce RTX 3090 GPU, with 24 MB of RAM\textemdash the same GPU used in ESSL experiments. Though we include results from large self-supervised methods as a reference, the selected baseline enables results comparison to a method proposed with sustainable goals in mind. 

\begin{table}[hbt]
\centering
\begin{tabular}{ p{3.5cm} m{1.5cm}  m{1.5cm} }
    \toprule
    Parameter  & ESSL & S3LSpeech  \\
    \midrule
    \midrule
    Training time (hrs)      & 28   & \textbf{19} \\ 
    Attention layers         & 12   & \textbf{4}  \\
    Trainable parameters (M) & 90.8 & \textbf{23.2} \\ 
    Model size (MB)          & 729  & \textbf{188} \\ 
    \bottomrule
    \end{tabular}
    \caption{Comparison between S3LSpeech and ESSL models. Model size includes both teacher and student encoders.}
    \label{tab:model_comparison}
\end{table}
\vspace{-0.8cm}

\begin{table}[hbt]
\centering
\begin{tabular}{ p{6.0cm} m{1.0cm}  }
    \toprule
    SSL Model  & ASR  \\
    \midrule
    \midrule
    MelHuBERT \cite{lin2022melhubert}        & 11.3 \\
    wav2vec2 Base \cite{baevski2020wav2vec}  & 4.79  \\
    HuBERT Base \cite{hsu2021hubert}         & 4.79  \\
    Spiral Base \cite{huang2022spiral}       & 3.30  \\
    WavLM Base  \cite{chen2022wavlm}         & 3.40  \\
    ComSSL \cite{zhang2020pushing}      & \textbf{1.40}  \\
    \midrule
    ESSL                                     & 10.687 \\
    S3LSpeech                                & \textbf{7.997}  \\ 
    \bottomrule
    \end{tabular}
    \caption{WER performance for the LibriSpeech test-clean dataset. Differences against baseline results have $p \leq 0.05$ for the Student's t-test.} 
    \label{tab:large_model_results}
\end{table}
\vspace{-0.8cm}

\begin{table}[hbt]
\centering
\begin{tabular}{ p{3.0cm} m{1.5cm}  m{1.5cm} }
    \toprule
    Dataset      & ESSL & S3LSpeech  \\
    \midrule
    \midrule
    dev-other    & 28.181 & \textbf{23.345} \\ 
    dev-clean    & 10.378 & \textbf{7.628}  \\
    test-other   & 29.982 & \textbf{24.639} \\ 
    test-clean   & 10.687 & \textbf{7.997}  \\ 
    \bottomrule
    \end{tabular}
    \caption{S3LSpeech improvements in ASR against the baseline method. Results include WER performance with a language model on LibriSpeech datasets. Differences against baseline results have $p \leq 0.05$ for the Student's t-test.}
    \label{tab:model_comparison_librispeech}
\end{table}
\vspace{-0.8cm}

\subsection{Results and discussion}

\begin{figure}[htb]
\centering
\includegraphics[width=0.75\columnwidth]{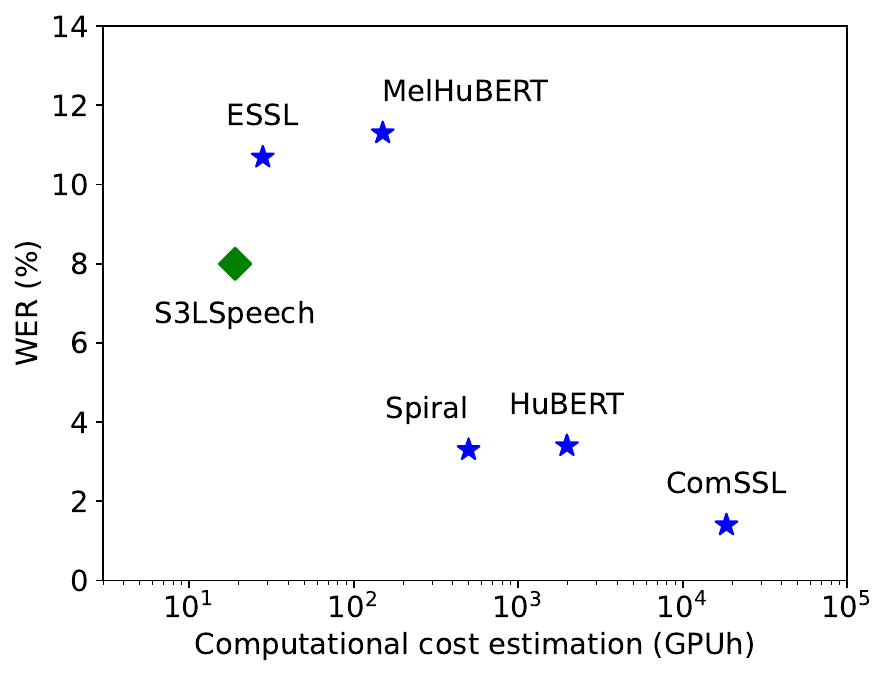}
\caption{Comparison between WER for Librispeech test-clean dataset and computing cost estimation. The X-axis has a logarithmic scale.}
\label{fig:s3lspeech_computational_cost}
\end{figure}

\begin{figure}[htb]
\centering
\includegraphics[width=0.80\columnwidth]{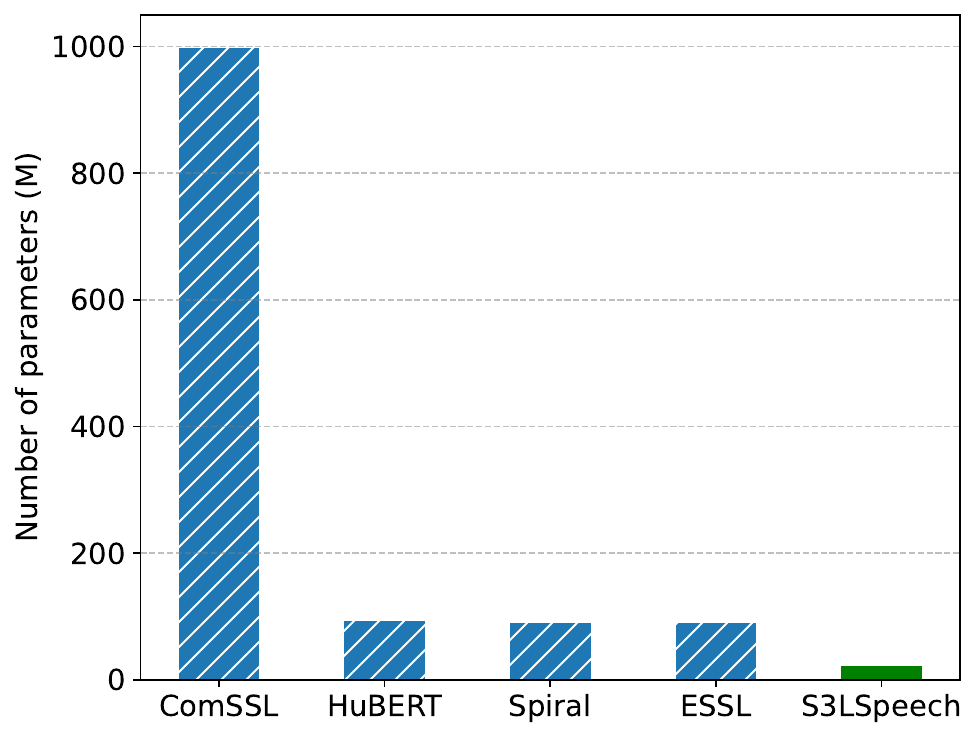}
\caption{Memory usage improvement of S3LSpeech against existing self-supervised learning models.}
\label{fig:s3lspeech_memory_usage}
\end{figure}

S3LSpeech outperforms the baseline in several aspects, including memory usage, word error rate (WER), and computing cost estimations (Tables \ref{tab:model_comparison} and \ref{fig:s3lspeech_memory_usage}). Pretraining takes only 19GPUh, while the baseline needs 28GPUh, representing a 30\% improvement in computing costs (Figure \ref{fig:s3lspeech_computational_cost}). Similarly, WER decreases from 10.687\% to 7.997\% in the Librispeech test-clean dataset. In terms of memory usage, encoder's trainable parameters decrease from 90.8M to 23.2M, representing a 75\% improvement. 

Concerning batch sizes, using a single GPU can severely limit the amount of audio in each batch. Dynamic batching and gradient accumulation enables to deal with this issue. The GPU supports a batch size of 18 minutes with dynamic batching, while training with mixed precision. After four gradient accumulations, it gets to 72 minutes, without generating Out of Memory (OOM) exceptions (Table \ref{tab:model_configuration}). This batch size is close to batch sizes in large speech models: DinoSR uses 63 minutes \cite{liu2023dinosr}, HuBERT uses 47 minutes \cite{hsu2021hubert}, WavLM uses 187 minutes \cite{chen2022wavlm}, and wav2vec2 uses 96 minutes \cite{baevski2020wav2vec}.

Another advantage of S3LSpeech is the possibility to train the speech representations in mobile devices or edge platforms. Most mobile devices have a single GPU in the System-on-Chip, while edge platforms like the NVidia Jetson Xavier series, have a single GPU \cite{gaol2023match}. Additionally, it is possible to pretrain the model using federated self-supervised learning on edge platforms. In federated learning, data is not centralized, which requires that each client machine train with the data available to it \cite{kairouz2021advances}. All client machines participate in the training, so having a method with low computing costs and minimal memory usage is crucial, particularly in applications where client machines are mobile devices or edge platforms \cite{gaol2023match}.

\begin{table}[hbt]
\centering
\begin{tabular}{ p{3.5cm} m{1.5cm}  m{1.5cm} }
    \toprule
    Configuration  & dev-other & dev-clean  \\
    \midrule
    \midrule
    S3LSpeech                 & \textbf{23.345} & \textbf{7.628} \\ 
    - Mamba                   & 30.379 & 10.718 \\
    - no positional shifting  & 26.733 & 12.065 \\
    - no audio augmentations  & 35.192 & 14.644 \\
    - no pretraining          & 98.965 & 99.077 \\
    - no dynamic batching     & OOM    & OOM    \\
    \bottomrule
    \end{tabular}
    \caption{Analysis of different configurations for the proposed model. Results include WER performance on LibriSpeech dev-other and dev-clean datasets. Out of Memory (OOM) exceptions appear when removing dynamic batching.} 
    \label{tab:model_configuration}
\end{table}
\vspace{-0.6cm}

Despite previously mentioned advantages, S3LSpeech has a limitation in very-low data settings. Indeed, very-low data settings are challenging, particularly for under-resourced languages \cite{lugo2024towards}. S3LSpeech generates a 95.48\% WER in the LibriSpeech dev-clean dataset when finetuning with Librilight 10 min, 95.74\% WER with Librilight 1 hr, and 96.29\% WER with Librilight 10 hr, indicating that S3LSpeech fails to properly converge in very-low data settings.

We also did several experiments to analyze the S3LSpeech approach (Table \ref{tab:model_configuration}). Using Mamba\textemdash a recently proposed layer that relies on structured state space models to replace self-attention layers \cite{gu2023mamba}\textemdash degrades model performance, as WER decreases to 10.718\% in the Librispeech dev-clean dataset. This degradation happens despite doubling the number of Mamba layers to match trainable parameter count. A similar scenario happens when removing SpecAugment and noise, highlighting the role of audio augmentations for self-supervised training. Performance degrades also when removing random positional shifting, suggesting the importance of forcing the model to learn exclusively from input audio by discarding positional information. Besides, finetuning from random weights is not enough to train the model. Without pretraining, the model fails to converge. Regarding dynamic batching, it is essential for training, removing it impedes training because of OOM exceptions. 

Lastly, when comparing S3LSpeech against large self-supervised models, there are substantial improvements in memory usage and computing costs (Figure \ref{fig:s3lspeech_computational_cost} and \ref{fig:s3lspeech_memory_usage}). There is, however, a reduction in ASR performance (Table \ref{tab:large_model_results}), but improvements remain substantial. For instance, when comparing S3LSpeech against SPIRAL, trainable parameter reduction represents a 75\% improvement, as trainable parameters go from 95M to 23M. Regarding computing costs, there is an order of magnitude improvement, as costs go from 499 GPUh to 19 GPUh. Similarly, when comparing against ComSSL, trainable parameter reduction represents a 97\% improvement, going from 1000M to only 23M trainable parameters. In terms of computing costs estimation, costs go from 18432 GPUh to only 19 GPUh. This improvement represents almost three orders of magnitude reduction in computing cost estimations.                      

\section{Conclusion}
\label{sec:conclusion}

In sum, we proposed S3LSpeech, a sustainable self-supervised model for learning speech representations, taking into account environmental concerns from high computing costs. S3LSpeech comprises a teacher\textendash student architecture for self-supervised pretraining, which takes less than 24 hours with a single GPU. After pretraining, it is possible to finetune the model for downstream tasks. Tests with ASR as a downstream task show improvements in error rate against a resource-constrained baseline. At the same time, pretraining cost reductions represent a 30\% improvement in computing cost estimations and a 75\% reduction in memory usage against the baseline. A comparison to large speech models represents a substantial improvement, having a 97\% reduction in memory usage and almost three orders of magnitude reduction in computing cost estimations. In general, our work is part of the quest to develop environmentally responsible models, striving for the sustainable development of artificial intelligence.

In future work, we will extend the downstream evaluation of speech representations to other downstream tasks\textemdash keyword spotting, speaker identification, emotion recognition, query by example spoken term detection, slot filling, automatic speaker verification, and speaker diarization. We will also work in the data efficiency direction, investigating model convergence in very-low data configurations. 

\bibliographystyle{IEEEtran}
\bibliography{references}

\end{document}